\documentclass[conference,letterpaper]{IEEEtran}
\IEEEoverridecommandlockouts
\makeatletter
\newcommand{\linebreakand}{%
  \end{@IEEEauthorhalign}
  \hfill\mbox{}\par
  \mbox{}\hfill\begin{@IEEEauthorhalign}
}
\makeatother
\usepackage[caption=false,font=normalsize,labelfont=sf,textfont=sf]{subfig}
\usepackage{makecell}
\usepackage{cite}
\usepackage{amsmath,amssymb,amsfonts}
\newtheorem{definition}{Definition}
\usepackage{algorithmic}
\usepackage{graphicx}
\usepackage{textcomp}
\usepackage{xcolor}
\usepackage{threeparttable}
\def\BibTeX{{\rm B\kern-.05em{\sc i\kern-.025em b}\kern-.08em
    T\kern-.1667em\lower.7ex\hbox{E}\kern-.125emX}}
\begin{document}

\title{Seeking to Collide: Online Safety-Critical Scenario Generation for Autonomous Driving with Retrieval Augmented Large Language Models \\
\thanks{*Research supported by the National Natural Science Foundation of China under Grants [524B2164, 52125208, 52422215].

$^\dagger$Corresponding author: Tong Nie (tong.nie@connect.polyu.hk).
}}

\author{\IEEEauthorblockN{Yuewen Mei}
\IEEEauthorblockA{\textit{Dept. of Traf. Eng.} \\
\textit{Tongji University}\\
Shanghai, China \\
meiyuewen@tongji.edu.cn}
\and
\IEEEauthorblockN{Tong Nie}
\IEEEauthorblockA{\textit{Dept. of Civ. \& Envir. Eng.} \\
\textit{The Hong Kong Polytechnic University}\\
Hong Kong SAR, China \\
tong.nie@connect.polyu.hk}
\and
\IEEEauthorblockN{Jian Sun}
\IEEEauthorblockA{\textit{Dept. of Traf. Eng.} \\
\textit{Tongji University}\\
Shanghai, China \\
sunjian@tongji.edu.cn}
\and
\IEEEauthorblockN{Ye Tian}
\IEEEauthorblockA{\textit{Dept. of Traf. Eng.} \\
\textit{Tongji University}\\
Shanghai, China \\
tianye@tongji.edu.cn}
}

\maketitle

\begin{abstract}
Simulation-based testing is crucial for validating autonomous vehicles (AVs), yet existing scenario generation methods either overfit to common driving patterns or operate in an offline, non-interactive manner that fails to expose rare, safety-critical corner cases. In this paper, we introduce an online, retrieval-augmented Large Language Models (LLMs) framework for generating safety-critical driving scenarios. Our method first employs an LLM-based behavior analyzer to infer the most dangerous intent of the background vehicle from the observed state, then queries additional LLM agents to synthesize feasible adversarial trajectories. To mitigate catastrophic forgetting and accelerate adaptation, we augment the framework with a dynamic memorization and retrieval bank of intent-planner pairs, automatically expanding its behavioral library when novel intents arise. Evaluations using the Waymo Open Motion Dataset demonstrate that our model reduces the mean minimum time-to-collision from 1.62 to 1.08 s and incurs a 75\% collision rate, substantially outperforming baselines.
\end{abstract}


\section{Introduction}
Simulations are indispensable for developing and testing autonomous vehicles (AVs), as real-world on-road testing cannot feasibly encounter the vast space of possible driving conditions.
A central challenge is to generate realistic and diverse traffic scenarios that capture complex multi-agent interactions. Early simulators used rule-based or handcrafted models, but these lack human-like variability \cite{chen2024data}. Data-driven methods such as the imitation learning (IL) from driving logs \cite{feng2023trafficgen} or generative models \cite{pronovost2023scenario} have become popular to model agent behaviors more faithfully.
However, such approaches often overfit the naturalistic driving distribution: they tend to reproduce common traffic patterns while rare or adversarial events are underrepresented. For example, IL methods typically learn a policy to reconstruct the realistic behaviors using ground truth trajectories as supervision. Likewise, generative techniques such as GANs or diffusion can synthesize new scenarios, but their outputs remain anchored to the training distribution \cite{ding2023survey}. In practice, this means safety-critical corner cases, i.e., the long tail of rare events are often overlooked, leaving AVs untested on precisely the situations they must eventually handle.

\begin{figure}[htbp!]
    \centering
    \includegraphics[width=0.99\linewidth]{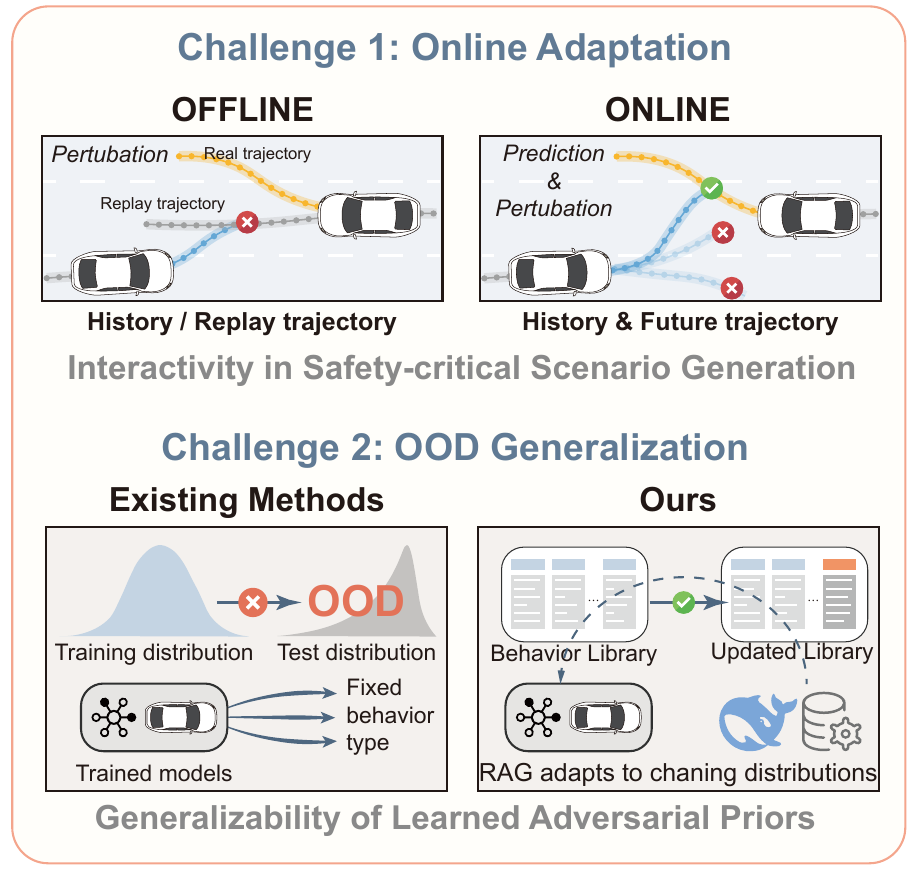}
    \caption{\textbf{Challenges of interactive safety-critical scenario generation.} This paper highlights two basic challenges in safety-critical scenario generation: (1) online adaptation, where adversarial priors must interact with an agent’s evolving behaviors in the future, and (2) out-of-distribution (OOD) generalization, where learned priors must remain effective under novel test distributions. Existing offline methods only perturb historical trajectories under a fixed behavior distribution. Our proposed framework generates scenarios in online and continually updates safety-critical behavior library using LLMs.}
    \label{fig:intro}
\end{figure}

To alleviate the ``curse of rarity'' issue \cite{liu2024curse} of safety-critical scenario generation, recent studies have developed adversarial methods to intentionally generate such scenarios \cite{ding2020learning,wang2021advsim,rempe2022generating,zhang2023cat,hao2023adversarial,mei2024bayesian,stoler2024seal,mei2025llm}. 
However, existing approaches largely focus on complex optimization processes, which are often computationally intensive and static by design. Methods based on adversarial learning, search-based perturbations, or risk-driven behavior modeling have shown promise in constructing challenging scenarios; however, they typically require pretraining and rely on fixed objectives or hand-crafted heuristics. This limits their flexibility to continuously adapt to the evolving behaviors of autonomous vehicles in real-world driving environments.

Recent advances in large language models (LLMs) offer a new paradigm for scenario generation \cite{chang2024llmscenario,nie2025exploring}, especially for the scenario of interests. By interpreting rich natural language prompts and world knowledge, LLMs can flexibly compose traffic scenarios beyond the training set. For example, ChatScene \cite{zhang2024chatscene} leverages an LLM with a knowledge retrieval module to translate textual scenario descriptions into domain-specific simulator code that reproduces rare events. These LLM-driven methods can produce diverse and even safety-critical scenarios, guided by human language instruction. 

In summary, despite these advances, existing approaches still have two significant limitations, as illustrated by Fig. \ref{fig:intro}.
First, most scenario generators, whether IL-based or LLM-based, operate \textbf{offline}: they generate a fixed set of scenarios to play out, without adapting to the AV’s real-time behavior. Without the ability to interact in online, these generators risk missing unexpected failure modes, ultimately reducing the effectiveness of safety validation \cite{ding2023survey}. 
Second, although some works have explored online methods (e.g. reinforcement learning to edit scenario parameters or introducing a behavior model \cite{zhang2023cat,ransiek2024goose,mei2025llm}), they largely obtain adversarial behavioral priors as a separate offline process, e.g., a learned distribution or a prescribed behavior dictionary. As a result, current methods can face difficulty in out-of-distribution (\textbf{OOD}) generalization to continuously tune and adapt new scenarios during testing. 
Addressing these gaps requires a new approach that can generate high-risk scenarios online, dynamically adjusting to evolving behaviors to more rigorously stress test robustness. 

While current LLM-based scenario generators only close the loop in limited ways. The challenge remains to integrate LLMs flexibly into online and interactive scenario generation. To this end, this paper identifies the opportunity for an online and safety-critical scenario generation method. 
Specifically, we propose an agent framework by exploiting the power of retrieval-augmented LLMs. By analyzing behavioral intent in historical scenes, an initial LLM first output the dangerous behavior type and required parameters. These conditions are used to prompt other LLM agents to generate and refine feasible future adversarial trajectories, thus forming an interactive and safety-critical scenario. To further adapt the model to both familiar and novel scenarios during testing, a dynamic memorization and retrieval mechanism is developed.

Our contributions are summarized as follows:
\begin{enumerate}
    \item An LLM-based agent framework is proposed to generate interactive and safety-critical scenarios online.
    \item A memorization and retrieval mechanism is developed to continuously adapt LLMs to changing scenarios.
    \item Experiments using the Waymo Open Motion Dataset show the effectiveness of both adversarial trajectory generation and memorization-retrieval methods.
\end{enumerate}

The remainder of this paper is organized as follows. 
Section \ref{sec:related work} reviews related work.
Section \ref{sec:methodology} formulates the methodology.
Section \ref{sec:experiment} conducts experiments to verify effectiveness of the proposed methodology. Section \ref{sec:conclusion} concludes the work.

\section{Related Work}\label{sec:related work}

\subsection{Data-Driven Realistic Traffic Scenario Generation}
A large amount of work has focused on learning driving behaviors from data. Imitation learning (behavior cloning, GAIL, etc.) uses naturalistic driving logs to train policies for traffic agents \cite{tan2021scenegen,choi2021trajgail,feng2023trafficgen}. These methods can capture human-like motion in isolation, but struggle with multi-agent coordination. Joint trajectory optimization or inverse reinforcement learning (IRL) can explicitly model interactions, but such methods often do not scale to general scenes and require complex cost design. Generative models have been applied to broaden scenario diversity. GANs and variational models can synthesize new trajectories or traffic scenes. For instance, conditional GANs have been used to extrapolate highway lane-change scenarios \cite{li2024vehicle}. However, GANs are computationally expensive and heavily dependent on training data. Diffusion models have also been proposed for traffic generation \cite{pronovost2023scenario,xu2025diffscene}. They can iteratively refine scenarios and allow controllable editing, but training such diffusion policies remains challenging and still anchors output to the training distribution. 
In particular, if corner cases are sparse or missing from the data, the learned generator will simply not produce them. 
In summary, data-driven generators tend to replicate the naturalistic distributions they are trained on, and thus rarely produce truly novel or dangerous events.

\subsection{Safety-Critical Scenario Generation
}
A parallel line of work focuses explicitly on generating rare, safety-critical scenarios for adversarial testing of AVs. 
To explicitly target high-risk cases, some methods formulate scenario generation as a search or optimization problem.
These ``corner-case" generators often use adversarial or search-based methods \cite{ding2020learning,wang2021advsim,rempe2022generating,zhang2023cat,hao2023adversarial,stoler2024seal,mei2025llm}. For example, Hao et al. \cite{hao2023adversarial} combine naturalistic driving models with a RL-based adversarial training loop. They calibrate nominal human-driving behaviors from real data and then fine-tune them to produce realistic adversarial scenarios. However, it still operates offline: the RL agent is trained once and then generates scenarios for later testing.
Other works formulate scenario generation as trajectory perturbation or influence maximization (e.g. adversarial agents \cite{zhang2023cat}, SEAL \cite{stoler2024seal}), emphasizing worst-case driving conditions. In particular, Zhang et al. \cite{zhang2023cat} present a closed-loop framework for adversarial behavior generation. They determine the behavior of adversarial agent by calculating the collide likelihood, then optimize agent parameters to reach those boundary conditions. 

Despite these advances, most safety-critical generators remain largely offline or computationally heavy. Many are formulated as optimizations or RL that require extensive training. They often rely on hand-tuned priors or optimization targets, which may still miss unforeseen extreme cases. Furthermore, few fully integrate with the AVs in online settings. 
However, AV testing should ideally be adaptive: the scenario should evolve in response to the behavior of the ego vehicle. Scripted or replay-based scenarios inherently cannot adapt, and even closed-form RL generators do not interact with the AV in real time. 
Non-interactive tests cannot fully expose AV weaknesses. In practice, there is a lack of online safety-critical scenario generation method that reacts continuously to the AV’s actions.

\subsection{LLM-Enhanced Scenario Generation}
Emerging work uses LLMs to facilitate scenario design. These methods leverage the world knowledge and flexible reasoning ability of LLMs to synthesize scenarios from high-level instructions or demonstartions. For example, LLMScenario \cite{chang2024llmscenario} designs chain-of-thought prompts to generate scenarios and score them for realism and rarity, explicitly targeting uncommon events. 
OmniTester \cite{lu2024multimodal} uses LLM prompts together with tools like SUMO to generate controllable test cases.
These methods highlight that LLMs can interpret user specifications and world knowledge to create diverse, structured scenarios.
Notably, some works integrate LLMs with knowledge retrieval frameworks to bridge language and simulation. ChatScene \cite{zhang2024chatscene} uses an LLM agent with a retrieval database of scenario-description/code pairs to first generate high-level adversarial scenario descriptions and then translate them into executable simulator code. RealGen \cite{ding2024realgen} retrieves similar traffic examples and in-context learns to compose new scenario layouts.
These LLM-based methods have broadened the scope of scenario generation. They can compose unseen scenario variations by mixing and matching examples or crafting new narratives, potentially mitigating the overfitting issue of pure IL approaches.

Nevertheless, almost all existing LLM-driven approaches still operate in a mostly offline fashion: the LLM produces a scenario (or code) which is then run in simulation, but the scenario itself is not adapted in the loop. Recent work by Mei et al. \cite{mei2025llm} begins to close this loop. 
However, it is an early step in closed-loop generation and does not incorporate retrieval-augmented knowledge.

\subsection{Summary of Challenges}
Across these domains, several common limitations emerge. (a) Overfitting to naturalistic data: Generators trained on real driving data reproduce common scenarios but rarely sample the rare corner cases needed for safety assurance. (b) Offline generation: Both traditional and LLM-based methods typically generate scenarios in advance. Rare exceptions aside, they do not adapt scenarios during a simulation run, which means that the AV is not tested in a truly interactive manner. (c) Lack of closed-loop interaction: There is no unified framework to dynamically adjust scenario parameters based on the AV’s real-time behavior. Existing adversarial and adaptive methods either require handcrafting or heavy computation, and do not leverage rich external knowledge. 


\section{Methodology}\label{sec:methodology}
In this section, we present an online, intent‐driven framework for the safety‐critical scenario generation. Our method proceeds in three stages: (1) safety-critical background vehicle behavior inference, where an LLM analyzes historical state to assign an intent label; (2) intent‐conditioned trajectory synthesis, in which an LLM‐based code agent produces a planner that generates dangerous trajectories according to the intent; and (3) dynamic memorization and retrieval, which augments the LLM with a structured cache of previously generated planners to mitigate forgetting and accelerate online operation. Together, these components enable adaptive generation of hazardous yet feasible background vehicle maneuvers, allowing us to convert safe driving histories into safety‐critical future scenarios online.

\begin{figure*}
    \centering
    \includegraphics[width=0.99\linewidth]{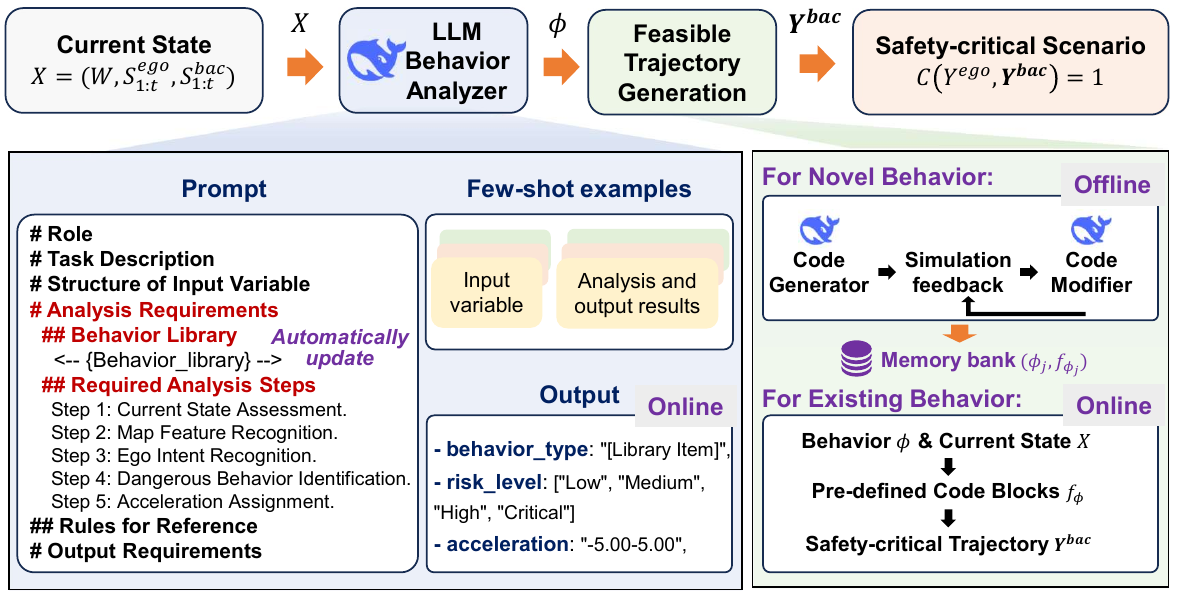}
    \caption{\textbf{Overview of the proposed LLM-driven framework for safety-critical scenario generation.} Given the current state \( X = (W, S_{1:t}^{ego}, S_{1:t}^{bac}) \), an LLM-based Behavior Analyzer uses structured prompts and few-shot examples to infer dangerous behavior types, risk levels, and acceleration profiles. These are used to generate feasible trajectories and incur safety-critical scenarios where \( C(Y^{ego}, Y^{bac}) = 1 \). Offline, code blocks for behavior types are generated and refined through simulation; online, these memories are matched with real-time behaviors and retrieved to augment the generation of similar scenarios.}
    \label{fig:pompt}
\end{figure*}

\subsection{Problem Formulation}
\begin{definition}[Traffic Scenario]
We consider an autonomous driving scenario at time $t$ defined by
\begin{equation}
X \;=\;\Bigl(W,\; S_{1:t}^{ego},\; \mathbf{S}_{1:t}^{bac}\Bigr),
\end{equation}
where $W$ denotes the static road geometry (lane center-lines, etc.), $S_{1:t}^{ego} = \bigl(s_1^{ego}, s_2^{ego}, \dots, s_t^{ego}\bigr)$ is the observed states of the \emph{ego} vehicle up to time $t$, $\mathbf{S}_{1:t}^{bac} = \{\,S_{1:t}^{i}\}_{i=1}^M$ collects the trajectories of the $M$ background vehicles up to time $t$.

Future trajectories from $t$ to horizon $T$ are denoted as:
\begin{equation}
Y^{ego} \;=\; S_{t:T}^{ego}, 
\quad
\mathbf{Y}^{bac} \;=\;\{\,Y^{i}\}_{i=1}^M,
\end{equation}
where $Y^{i} = S_{t:T}^{i}$.
\end{definition}

\begin{definition}[Safety‐Critical Scenario]
Given an observed scenario $X$, a complete scenario is formulated as: 
\begin{equation}
\tau \;=\;\bigl(X,\;Y^{ego},\;\mathbf{Y}^{bac}\bigr).
\end{equation}

We define $\tau$ as \emph{safety‐critical} if a collision occurs between the ego and at least one background vehicle in the future rollout:
\begin{equation}\label{def:safety-critical}
\begin{aligned}
C\bigl(Y^{ego},\mathbf{Y}^{bac}\bigr)
\;&=\;
\mathbb{I}\Bigl[\exists\,i,\,\exists\,\kappa\in\{t,\dots,T\}:\,
\|y_\kappa^{ego}-y_\kappa^i\|\le\epsilon\Bigr]
\; \\
& =\;
{1,}
\end{aligned}
\end{equation}
where $\|\cdot\|$ is the Euclidean distance, $\mathbb{I}$ is an indicative function, and $\epsilon$ the collision threshold.
\end{definition}

Equivalently, in a probabilistic view, the collision can be formed as:
\begin{equation}
\mathbb{P}\bigl(C=1\mid X,\,Y^{ego},\,\mathbf{Y}^{bac}\bigr).
\end{equation}

\begin{definition}[Perturbation-based Safety‐Critical Scenario Generation]
Given the current state $X$, the task is to find proper futures of the background vehicle $\widehat{\mathbf{Y}}^{bac}$ that induce a safety-critical event according to the definition in Eq. \ref{def:safety-critical}.
\begin{equation}
\widehat{\mathbf{Y}}^{bac}
\;=\;
\arg\max_{\mathbf{Y}^{bac}}
\;\mathbb{P}\Bigl(C=1\;\big|\;X,\;Y^{ego},\;\mathbf{Y}^{bac}\Bigr).
\end{equation}

\end{definition}

\subsection{Overall Framework}
Based on above formulation, we elaborate our safety-critical scenario generation method that exploits the generative power of LLMs and an explicit memorization and retrieval process.
Directly prompting LLMs to generate raw trajectories can be challenging, as LLMs are trained in textual corpus and less effective at dealing with numeric data.
In our approach, we further decompose the generation of $\widehat{\mathbf{Y}}^{bac}$ into two stages:

\subsubsection{Behavioral‐Intent Inference}  We use an LLM‐based behavior analyzer agent $g_{\theta}$ to analyze the observed history into a discrete intent label for the background vehicles:
\begin{equation}
\phi \;=\; g_{\theta}\bigl(X),
\quad
\phi\in\Psi,
\end{equation}
where $\Psi$ is the library of dangerous behavioral‐intent types.

\subsubsection{Safety-critical Trajectory Generation}
Conditioned on the intent $\phi$ and current states $X$, we then generate the safety-critical future trajectories:
\begin{equation}\label{eq:traj_gen}
\widehat{\mathbf{Y}}^{bac}
\;=\;
f_{\omega}\bigl(\phi,\;X\bigr),
\end{equation}
where $f_{\omega}$ denotes trajectory generator conditioned on $\phi$, which is chosen to maximize the collision likelihood with the ego vehicle.

Overall, our online objective can be written as:
\begin{equation}\label{eq:overall-generation}
\max_{\theta,\,\omega}
\;\;
\mathbb{P}\Bigl(
C=1
\;\big|\;
X,\;
Y^{ego},\;
f_{\omega}\bigl(g_{\theta}(X),\,X\bigr)
\Bigr).
\end{equation}
  

In this way, normal histories are transformed into safety‐critical scenarios through intent‐guided trajectory synthesis. Eq. \ref{eq:overall-generation} can be extended to an iterative manner to dynamically interact with the ego vehicle, which can build a closed-loop testing environment \cite{mei2025llm}. We will detail the implementation of each component in the following.

\subsection{LLM-based Analyzer for Safety-critical Behavioral Intent}
To automatically identify dangerous behaviors of background vehicles that may threaten the ego vehicle under test in the scenarios, we design an LLM-based behavior analyzer. This module leverages LLMs with carefully engineered prompt and reasoning process to output the most safety-critical behavioral intent in the given scenario. 

As shown in Fig. \ref{fig:pompt}, the prompt is structured into six blocks: \textbf{Role}, \textbf{Task Description}, \textbf{Structure of Input Variables}, \textbf{Analysis Requirements}, \textbf{Rules for Reference} and \textbf{Output Requirements}. \textbf{Role} and \textbf{Task Description} assign the LLM the role of a safety behavior analyzer and instruct the model to predict behavioral intent in the system prompt. 

\textbf{Structure of Input Variables} clarifies the input format, including state features of both ego and surrounding vehicles and road layout. 
Recent studies suggest that the spatial reasoning and recognition capabilities of LLMs are limited \cite{chen2024spatialvlm}.
To help them handle arbitrary map geometries in a unified frame, we transform all positions into the ego‐centered frame. The transform of the coordinate denoted as:
\begin{equation}
\tilde s
\;=\;
R\bigl(-\theta_{t=0}^{ego}\bigr)\bigl(s - s_{t=0}^{ego}\bigr),
\end{equation}
where $s\in\mathbb{R}^2$ is a global 2D point, $s_{t=0}^{ego}$ and $\theta_{t=0}^{ego}$ are the ego’s initial position and heading, and $R(\alpha)$ is the 2D rotation matrix by angle $\alpha$.

\textbf{Analysis Requirements} includes the behavior library and the five required analysis steps. The behavior library is extensible and can be updated by LLM analyzer automatically, which is described in Section \ref{sec:expansion}.
Drawing inspiration from the chain-of-thought prompting method \cite{wei2022chain}, the LLM is guided through the five-step reasoning process. It first interprets the current states of both vehicles, then contextualizes these within the map topology. Based on the ego vehicle’s inferred intent, the model evaluate which behavior from the library would most likely result in a high-risk interaction. Finally, it assigns an appropriate acceleration $y_{acc}$ to simulate this behavior in downstream scenario generation. 

\subsection{Feasible Trajectory Generation with Behavioral Priors}
We now explain the implementation of Eq. \ref{eq:traj_gen}.
To synthesize realistic and threatening motion patterns of background vehicles, we propose a three-step trajectory generation pipeline guided by behavior-level priors.

\subsubsection{Hazardous Endpoint Inference}
Given an inferred intent label $\phi$, we aim to infer the background vehicle's likely future endpoint under this behavior. The LLM-based code generator is employed to produce executable code block $f_{\phi}$ that computes the endpoint coordinates $y^{bac}_T$ based on the current vehicle state and behavior semantics:

\begin{equation}
    y^{bac}_T = f_{\phi}(y^{bac}_t, y_{acc},T).
\end{equation}


\subsubsection{Interpolated Trajectory Planning}
To generate a target trajectory, we first invoke an LLM‐based code generator to produce a specialized trajectory planner:
\begin{equation}
\mathcal{C}(\psi)
\;\mapsto\;
f_{traj}\;=\texttt{Planner}_{\psi}(\,\cdot\,),
\end{equation}
where $\mathcal{C}\!:\Psi\to\mathcal{C}$ maps the user instruction to a block of executable planner code, and $\psi$ specifies the desired property and constraint of the planner.


Finally, the planner $f_{traj}$ (e.g.\ a multi‐display spline or polynomial interpolator) synthesizes a temporally smooth and kinematically feasible trajectory, which connects the start point $y^{bac}_t$ and the inferred endpoint $y^{bac}_T$:
\begin{equation}
\widehat{\mathbf{Y}}^{bac}
\;=\;
\bigl(y_{t+1}^{bac}, y_{t+2}^{bac},\dots,y_T^{bac}\bigr)
\;=\;
f_{traj}\bigl(y^{bac}_t, y^{bac}_T\bigl).
\end{equation}



\subsubsection{Simulation-Guided Code Refinement}
The generated trajectory $\widehat{\mathbf{Y}}^{bac}$ is further evaluated in the simulation testing of the ego vehicle. If the synthesized trajectory fails to induce meaningful safety-critical events, it is sent back to the Code Modifier, a refinement module that adjusts the original code block $f_{\phi}$ based on the feedback. This refinement continues until the synthesized trajectory meets the predefined risk threshold, ensuring the resulting scenario is genuinely adversarial and behaviorally consistent. The LLM-based code generator and the modifier are both implemented by prompting a general-purpose LLM with carefully engineered instructions.

\subsection{Online Scenario Generation by Dynamic Memorization and Retrieval}\label{sec:expansion}
Since online scenario generation is considered in this paper, the employed LLM agents such as behavior identifier and trajectory generator should quickly generate scenes similar to those previously encountered. However, due to the catastrophic forgetting of the LLM when the context is very long, additional memorization and retrieval mechanisms are needed to solve this problem. Therefore, we designed a dynamic memorization and retrieval mechanism that allows LLM to categorize the types of previously encountered scenarios and form a structured memory of the generated trajectory generator code. When encountering new scenarios, if LLM recognizes them as similar behavioral intents, then it will retrieve them based on the intent labels, and directly adopt the previously generated trajectory generators, thus speeding up the reasoning process.

To overcome long‐context forgetting in LLMs and accelerate online generation, we maintain a growing memory bank:
\begin{equation}
\mathcal{M}
\;=\;
\bigl\{\,(\phi_j,\;f_{\phi_j})\bigr\}_{j=1}^K,
\end{equation}
where each entry pairs a behavioral intent label $\phi_j\in\Psi$ with its previously generated planner $f_{\phi_j}$, and $K$ is the number of processed scenarios in the history. Given such a memory bank, the generation process can be augmented by retrieval:

\subsubsection{Intent‐based Retrieval}  At time $t$, given a newly inferred intent $\phi_t'$, 
we search $\mathcal{M}$ for the closest stored label:
\begin{equation}
j^*
\;=\;
\arg\min_{1\le j\le K}
\;d\bigl(\phi_t',\phi_j\bigr),
\end{equation}
where $d(\cdot,\cdot)$ is the similarity metric used to determine whether the $\phi_t$ is semantically similar to any intent $\phi_j$. If
$d\bigl(\phi_t',\phi_{j^*}\bigr)\;\le\;\epsilon_{\mathrm{ret}}$, we then retrieve and reuse the existing planner:
\begin{equation}
\widehat{f}_t \;=\; f_{\phi_{j^*}}.
\end{equation}
This avoids regenerating code for intents already seen.

\subsubsection{Expansion via Novel Intent Discovery}
In addition, since LLMs can encounter previously unseen scenarios, we also prompt LLMs to proactively explore new behavioral intent possibilities. If LLMs believe that the behavioral intent in the current scenario is not in the behavioral memory bank, LLMs need to understand and generate new behavioral intent labels, as well as generate the corresponding trajectory planner code and add it to the memory bank as a new memory. 

Formally, if no close match exists (i.e.\ $d(\phi_t,\phi_{j^*})>\epsilon_{\mathrm{ret}}$), the LLM Analyzer proposes a new intent label $\phi_{K+1}$, expanding the label set. Then, the corresponding planner $f_{\phi_{K+1}}$ is generated via the code‐generator agent. Finally, update memory:
\begin{equation}
    \mathcal{M}\;\leftarrow\;\mathcal{M}\;\cup\;\bigl\{\bigl(\phi_{K+1},\,f_{\phi_{K+1}}\bigr)\bigr\}, 
    \quad K\leftarrow K+1.
\end{equation}

Traditional adversarial generation methods, such as \cite{ding2020learning,wang2021advsim,xu2025diffscene}, generate safety-critical scenarios by fitting the training distributions. Once the model has been trained, the expected types of behavior to generate is fixed. This limits their effectiveness in dealing with dynamically changing scenarios, which is common in real-world driving environments. Instead, our method provides a flexible alternative to efficiently adapt the model to new scenarios.

\subsubsection{Full Online Loop} Combining retrieval and generation, for each scenario:
\begin{equation}
\widehat{f}_t
=
\begin{cases}
  f_{\phi_{j^*}}, 
    & d(\phi_t,\phi_{j^*}) \,\le\, \epsilon_{\mathrm{ret}},\\
  f_{\phi_{K+1}}, 
    & \text{otherwise (with memory update).}
\end{cases}
\end{equation}
The retrieved or newly generated planner $\widehat{f}_\phi$ is then invoked to produce the adversarial trajectory for the current background vehicle:
\begin{equation}
\widehat{\mathbf{Y}}^{bac}
\;=\;
f_{traj}\bigl(y_t^{bac}, \widehat{f}_\phi(y_t^{bac},y_{acc},X)\bigr).
\end{equation}

The generation of planner code blocks for each behavior type is an offline process. All verified behaviors and their corresponding endpoint computation code blocks are archived in $\mathcal{M}$ for online use.
This dynamic memorization and retrieval mechanism ensures that familiar intents are handled instantly by cached planners, while truly novel interactions prompt the LLM to extend its behavioral repertoire.

\section{Experiments}\label{sec:experiment}
This section presents the experimental setup and evaluation results of our proposed framework. We begin by describing the dataset and implementation details. Next, we evaluate the accuracy of the LLM-based behavior analyzer in identifying dangerous behaviors. Finally, we compare different scenario generation methods based on their ability to create high-risk situations.

\subsection{Datasets and Experimental Setups}
We conduct our experiments based on the Waymo Open Motion Dateset \cite{Ettinger_2021_ICCV_womd}. A total of 81 scenarios are imported, including 50 straight-road scenarios and 31 intersection scenarios. Each scenario contains an ego vehicle, an critical background vehicle, and multiple other surrounding vehicles. Our framework employs the DeepSeek-V3 and DeepSeek-R1 model, accessed via API.

\subsection{Safety-critical behaviors Identification}
LLM Behavior Analyzer is applied to identify the safety-critical behaviors of the background vehicle. Seven types of safety-critical behaviors are identified: Emergency Braking, Close Car-following, Aggressive Cut-in, Opposite Direction Intrusion, Intersection Rush-through Turn Left, Intersection Rush-through Go-straight, and Straight Lane Shift. We also manually annotated the most dangerous behavior of the critical background vehicle in each of the 81 scenarios, which serves as the ground truth for the LLM Behavior Analyzer.

\begin{figure}[t!]
    \centering
    \includegraphics[width=0.99\linewidth]{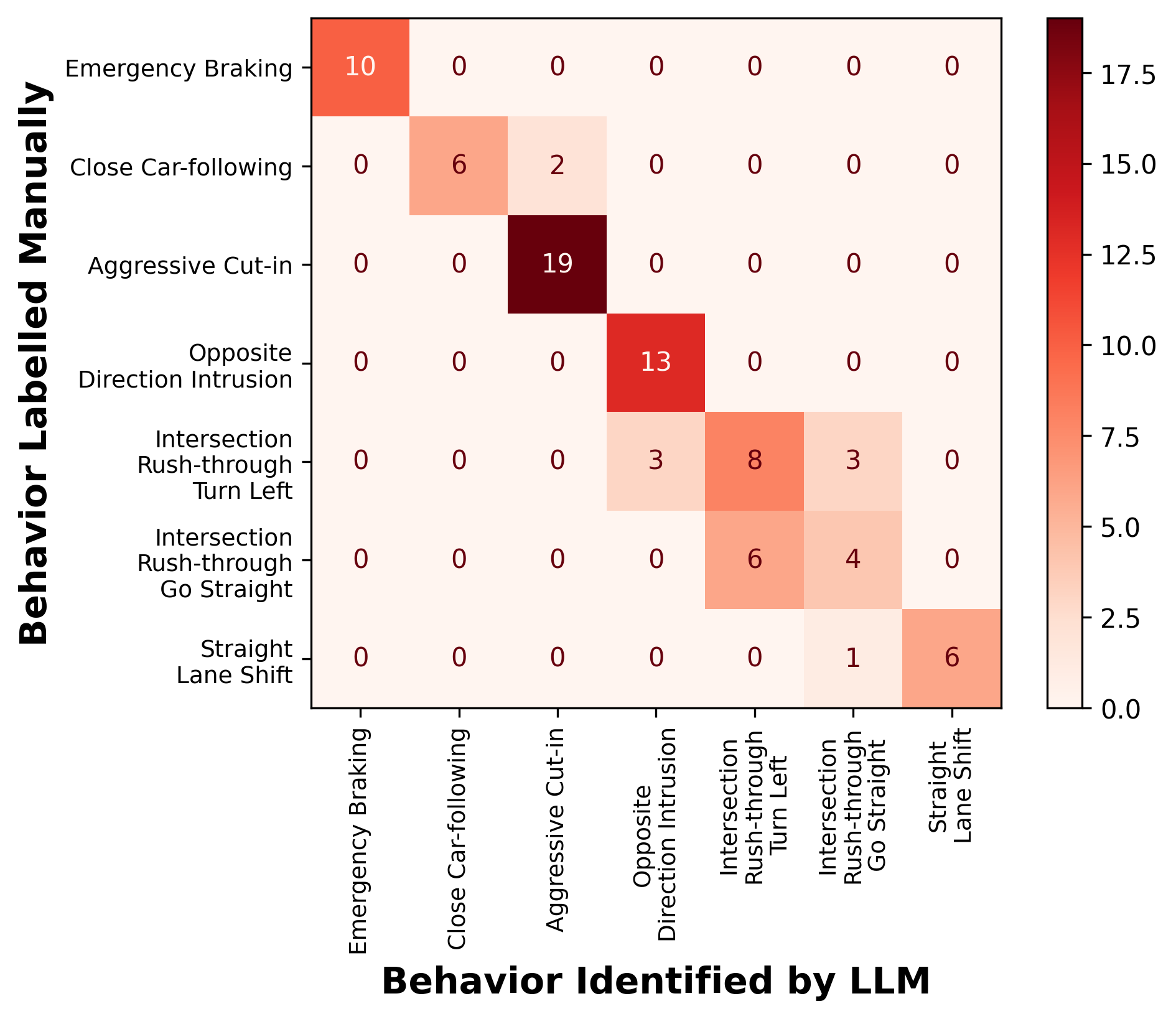}
    \caption{Confusion Matrix of Safety-Critical Behavior Identification by the LLM Behavior Analyzer. The LLM behavior analyzer achieves 81.5\% accuracy in identifying the most dangerous behavior of the critical background vehicle across 81 scenarios. The matrix reflects classification performance across seven behavior categories.}
    \label{fig:confusion_matrix}
\end{figure}

The confusion matrix of the classification results is shown in Fig. \ref{fig:confusion_matrix}. The overall accuracy of DeepSeek-R1 reached 81.5\%, demonstrating that the analyzer can reliably predict which background behavior would pose the highest risk to the ego vehicle. Meanwhile, the accuracy of DeepSeek-V3 is 74.1\%.

\begin{table}[t!]
    \centering
    \fontsize{9pt}{13pt}\selectfont
    \vspace{-10pt}
    \caption{Comparison of Safety-Critical Scenario Generation Methods}
    \begin{tabular}{c|cccc}
         \hline
         &MEAN Min TTC $\downarrow$	&Collision Rate	 $\uparrow$\\
         \hline
         Raw &1.62 &0.00 \\
         Transformer$^*$ &1.73 &0.41 \\
         \makecell{LLM-F (V3)}	&2.03 &0.22 \\
         \makecell{LLM-A (V3)} & 1.09 & 0.65 \\
         \makecell{LLM-A (R1)} & \textbf{1.08} & \textbf{0.75} \\
         \hline
    \end{tabular}
    \label{tab:safety_metrics}
    \begin{tablenotes}
    \footnotesize
    \item[1] $*$: Implemented by a Transformer-based candidate trajectory prediction model in \cite{Zhou_2022_CVPR}.
    \end{tablenotes}
\end{table}

\subsection{Safety-critical Scenario Generation}
After behavior identification, feasible future trajectories are calculated based on the code predefined by the llm code generator. To evaluate the safety-critical characteristics of the generated scenarios, we use two metrics: time to collision (TTC) and collision rate across the 81 scenarios. We report the mean minimum TTC and the proportion of scenarios that occur collision.

We compare the following three methods: 1) Transformer-based motion prediction model. A multi-modal motion prediction model \cite{Zhou_2022_CVPR} is used to generate candidate future trajectories for the background vehicle and the most dangerous one is selected. 2) LLM with fixed behavior library (LLM-F). The LLM can only select from a fixed set of two behaviors: Close Car-following and Emergency Braking. 3) LLM with auto-updated behavior library (LLM-A). The LLM selects and generates behaviors from the full and automatically updated behavior library.

The evaluation results are shown in TABLE \ref{tab:safety_metrics}. The LLM-A method with DeepSeek-R1 generates scenarios with a reduced mean minimum TTC from 1.62s to 1.08s, and causes collisions in 75\% of the scenarios, indicating a higher level of scenario criticality.

\begin{figure}[t!]
    \centering
    \subfloat[]{\includegraphics[width = 0.5\textwidth]{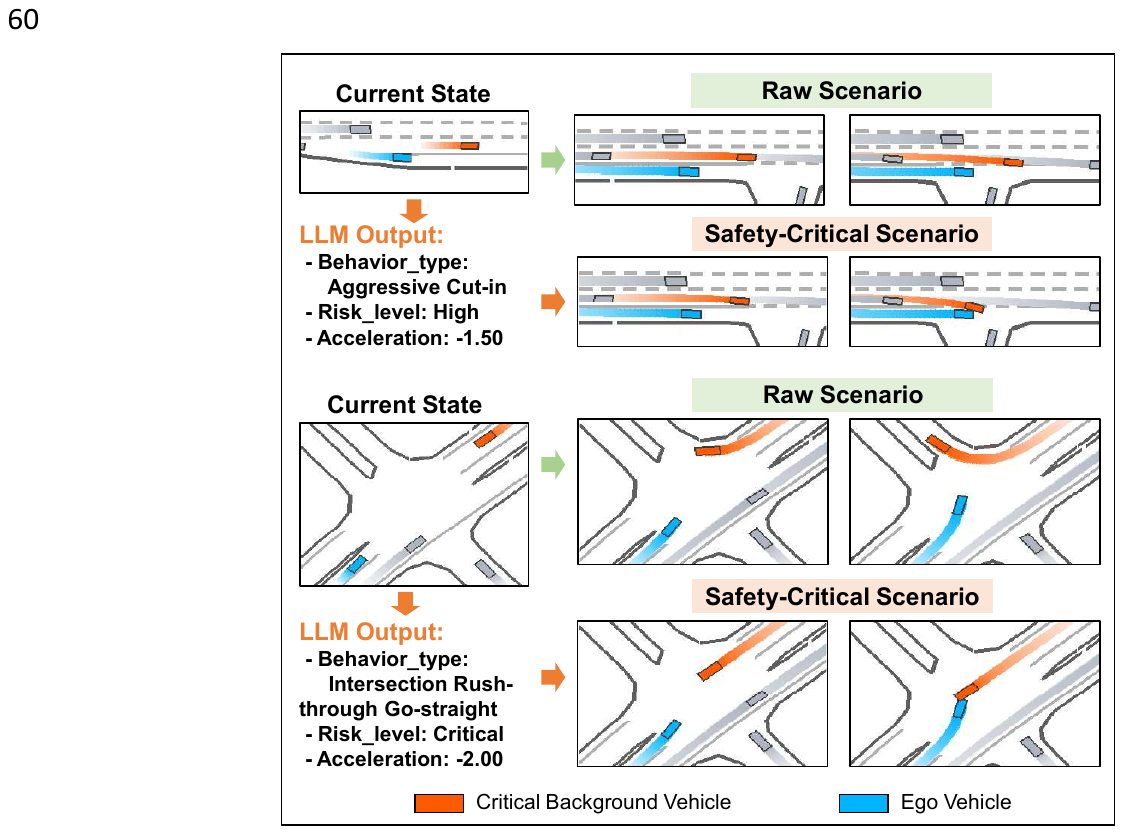}
    \label{fig:str_example}}
    \hfill
    \subfloat[]{\includegraphics[width = 0.5\textwidth]{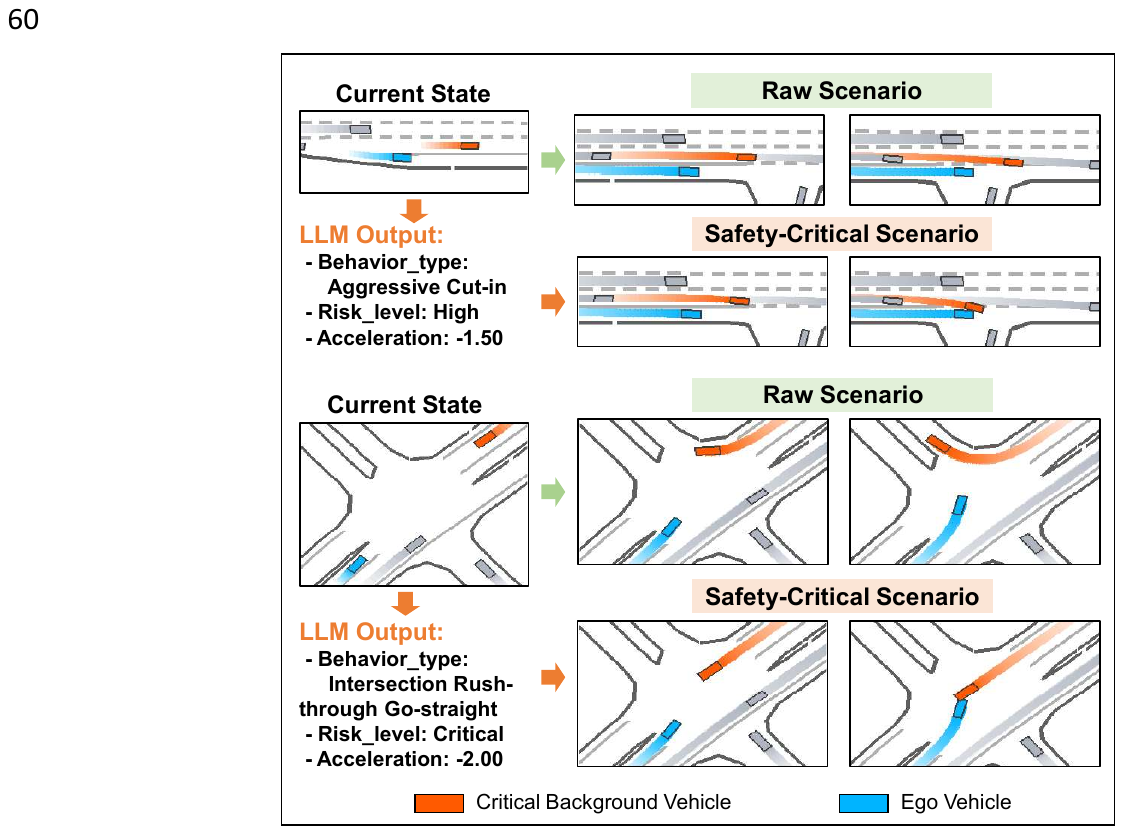}
    \label{fig:inter_example}}
    \caption{Examples of Generated Safety-Critical Scenarios. (a) In a straight-road scenario, the LLM identifies Aggressive Cut-in as the most dangerous behavior. The generated trajectory increases cut-in severity and leads to a collision.  (b) In an intersection scenario, the LLM modifies the background vehicle’s trajectory from a right turn to going straight, causing a conflict with the ego vehicle making a left turn.}
    \label{fig:example}
\end{figure}

\begin{figure}[t!]
    \centering
    \includegraphics[width=0.96\linewidth]{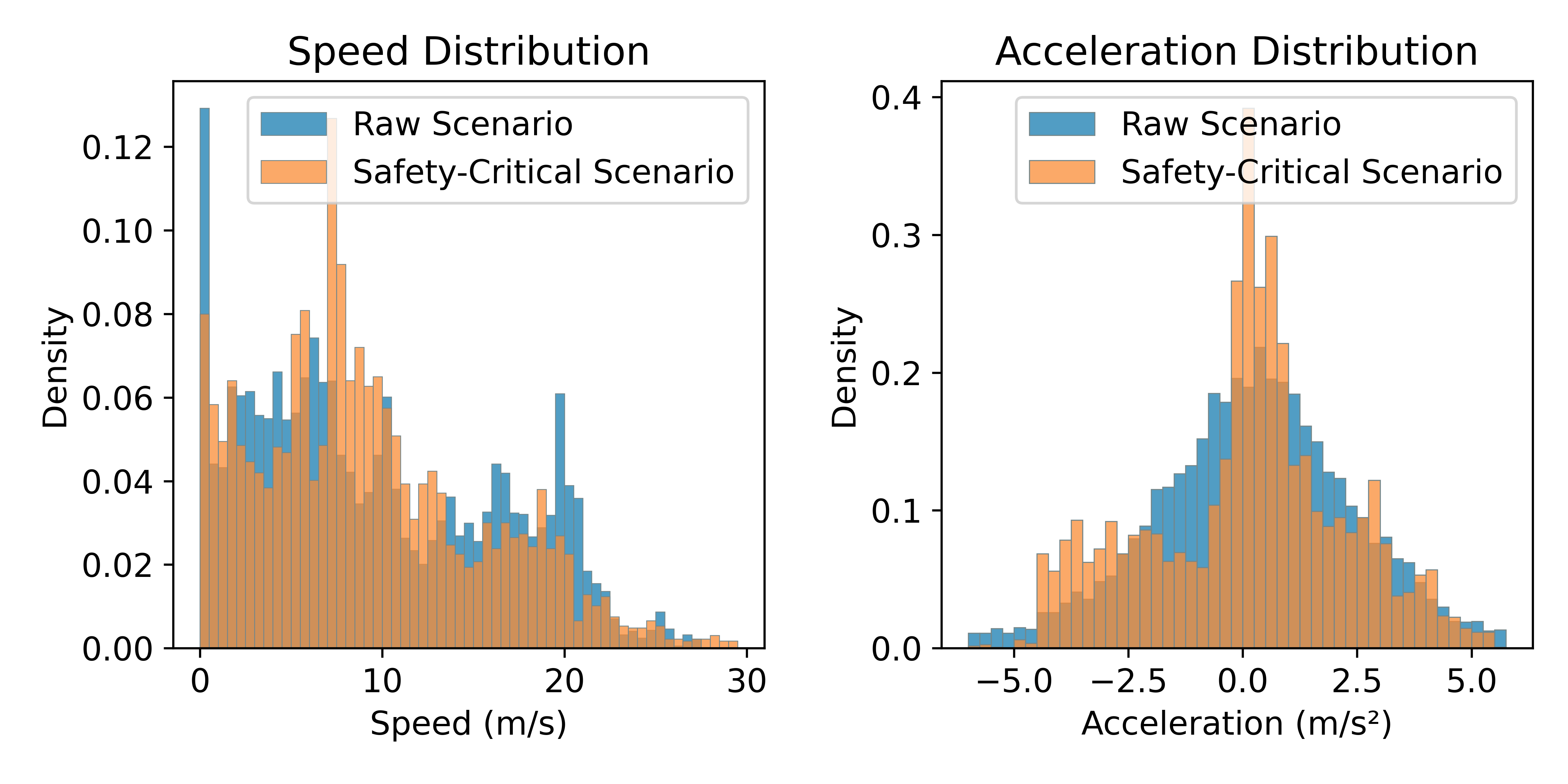}
    \caption{Comparison of speed and acceleration distributions between raw (blue) and generated safety-critical (orange) trajectories.}
    \label{fig:distribution}
\end{figure}

\begin{figure}[t!]
    \centering
    \includegraphics[width=0.86\linewidth]{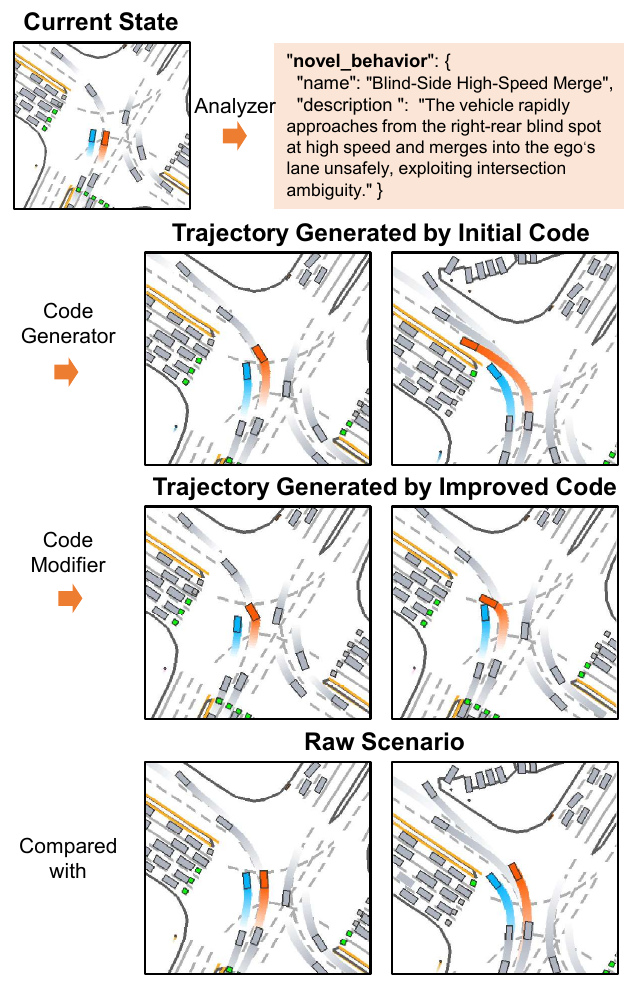}
    \caption{Discovery of a Novel Dangerous Behavior. The LLM identifies a new high-risk behavior involving a high-speed side merge during the dual left-turn scenario. The initial generated trajectory increases threat but avoids collision. After refinement by the code modifier, the trajectory leads to an intentional merge and collision.}
    \label{fig:new_example}
\end{figure}

Figure \ref{fig:example} shows two representative examples of generated scenarios. In Fig. \ref{fig:str_example}, the scenario is a straight road. The LLM identifies Aggressive Cut-in as the most dangerous behavior. The generated trajectory features a sharper and more sudden cut-in, leading to a collision with the ego vehicle, which fails to brake in time. In Fig. \ref{fig:inter_example}, the original intersection scenario has the ego vehicle turning left and the background vehicle making a right turn from the opposite direction. After modification, the background vehicle instead goes straight, resulting in a conflict with the turning ego vehicle.

To validate the realism of these generated scenarios, we compared their kinematic distributions with the raw data, as shown in Fig. \ref{fig:distribution}. The generated future trajectories remain statistically close to the original data, with low KL-Divergences for speed (0.141) and acceleration (0.138). The proportion of abnormal lateral acceleration ($\> 4 m/s^2$) is only 0.17\%. This demonstrates that while the generated scenarios exhibit more aggressive maneuvers to induce criticality, they adhere to physical constraints.

Additionally, Fig. \ref{fig:new_example} shows an example where the LLM Behavior Analyzer discovers a novel types of dangerous behavior not included in the initial behavior library. In this case, both vehicles are turning left. The LLM identifies that Blind-Side High-Speed Merge maneuver by the background vehicle would create danger. Then, the Code Generator outputs executable code to create a corresponding trajectory. Initially, the generated behavior increases the threat level but does not cause a collision. After refinement by the Code Modifier, the trajectory becomes more aggressive, and the background vehicle intentionally merges into the ego vehicle's path, leading to a collision.

\section{Conclusion}\label{sec:conclusion}
In this work, we present a novel framework for online safety-critical scenario generation that leverages retrieval augmented LLMs. By decomposing the task into behavioral-intent inference, trajectory synthesis, and memorization–retrieval mechanism, our method supports dynamically adaptation to evolving AV behaviors and novel driving contexts. Experiments demonstrate that our approach identifies the most dangerous background behaviors with over 80\% accuracy and produces scenarios with significantly higher collision rates compared to both data-driven and fixed-library baselines.

There are several directions for further enhancing interactive scenario generation. First, the uncertainty within the intent inference can be integrated. The current deterministic intent inference could be extended by leveraging the inherent stochasticity of LLMs. Second, extending the memory bank to support hierarchical intents and multi-agent coordination may yield more complex adversarial interactions. Third, while our current experiments are based on the Waymo dataset, the design of our framework allows for generalization, which should be extended to industry-standard simulators such as CARLA.

\section*{Declaration of generative AI and AI-assisted technologies in the writing process} During the preparation of this work, the authors used ChatGPT (an AI-assisted tool) to improve language clarity and readability. After using this tool, the authors reviewed and edited the content as needed and take full responsibility for the content of the published article.

\bibliographystyle{IEEEtran}
\bibliography{ref-extracts}

\end{document}